\documentclass{article} 
\usepackage{nips11submit_e,times}
\usepackage{cite}
\usepackage{tikz}

\usepackage{ upgreek }
\usepackage{verbatim} 

\usepackage{amssymb}
\usepackage{graphicx}
\usepackage{url}
\usepackage{verbatim} 
\usepackage{color}
\usepackage{wrapfig}

\usepackage[numbers]{natbib}
\usepackage{array}
\usepackage{relsize}
\usepackage{multirow}
\usepackage{amsmath}
\usepackage{hyperref}
\usepackage{rotating}

\usepackage[font=small,format=plain,labelfont={bf,up}]{caption,subfig}

\long\def\symbolfootnote[#1]#2{\begingroup%
\def\thefootnote{\fnsymbol{footnote}}\footnotetext[#1]{#2}\endgroup}
\newlength\savedwidth
\newcommand\whline[1]{\noalign{\global\savedwidth\arrayrulewidth
                               \global\arrayrulewidth #1} %
                      \hline
                      \noalign{\global\arrayrulewidth\savedwidth}}

\newenvironment{packed_enum}{
\begin{enumerate}
  \setlength{\itemsep}{1pt}
  \setlength{\parskip}{0pt}
  \setlength{\parsep}{0pt}
}
{\end{enumerate}}

\newenvironment{packed_item}{
\begin{itemize}
  \setlength{\itemsep}{1pt}
  \setlength{\parskip}{0pt}
  \setlength{\parsep}{0pt}
}{\end{itemize}}

\newlength{\sectionReduceTop}
\newlength{\sectionReduceBot}
\newlength{\subsectionReduceTop}
\newlength{\subsectionReduceBot}
\newlength{\abstractReduceTop}
\newlength{\abstractReduceBot}
\newlength{\captionReduceTop}
\newlength{\captionReduceBot}
\newlength{\subsubsectionReduceTop}
\newlength{\subsubsectionReduceBot}

\newlength{\horSkip}
\newlength{\verSkip}

\newlength{\figureHeight}
\title{3D Scene Grammar for Parsing RGB-D Pointclouds}
\author{Abhishek Anand and Sherwin Li\\
Department of Computer Science, Cornell University.\\
\texttt{\{aa755,sl855\}@cornell.edu}}

\nipsfinalcopy
\begin{document}
\maketitle


\begin{abstract}

We pose 3D scene-understanding as a problem of parsing in a grammar. 
A grammar helps us capture the compositional structure of real-word objects, e.g.,
a chair is composed of a seat, a back-rest and some legs.
Having multiple rules for an object helps us capture structural variations in objects, e.g., a 
chair can optionally also have arm-rests.
Finally, having rules to capture composition at different levels helps us formulate the 
entire scene-processing pipeline as a single problem of finding most likely 
parse-tree---small segments combine to form parts of objects, parts to objects and 
objects to a scene. We attach a generative probability model to our grammar by having a 
feature-dependent probability function for every rule.
We evaluated it by extracting labels for every segment and comparing the results 
with the state-of-the-art segment-labeling algorithm. Our algorithm was outperformed by the state-or-the-art method 
\footnote{ This paper was rejected at NIPS 2012. We thank our reviewers for their valuable comments. If you are interested in building on this work, you may contact the authors to get a copy of the reviews }.
But, Our model can be trained very efficiently (within seconds), and it scales only linearly in 
with the number of rules in the grammar. 
Also, we think that this is an important problem for the 3D vision community.
So, we are releasing our dataset \footnote{\url{https://www.dropbox.com/sh/ziocu03i3m932xq/Xwfn__0_qP}}
 and related code \footnote{\url{https://github.com/aa755/cfg3d}} .    





%

\end{abstract}

\section{Introduction}

With recent availability of inexpensive RGB-D sensors (e.g., Microsoft Kinect, that gives an RGB image together with depths, also called a `point-cloud'), several recent works have shown 
good
performance on several tasks, including segmentation \cite{pcl},
object recognition \cite{lai:icra11a,lai:icra11b}, object detection in scenes \cite{xiong:indoor,koppula2011semantic}, human activity detection \cite{sung:icra2012}, and so on. 
Most of these works label the 3D scene into a `flat' label-space, but there is no current
work that infers the deeper semantics by capturing the compositional 
structure of a 3D scene.

In this paper we present a method to parse the 3D scene (represented as a 3D point-cloud) using a probabilistic 3D scene grammar. To the best of our knowledge, our work is the first one to do so (see
Section~\ref{sec:related}).
Our grammar allows us to capture several properties, such as:

\begin{packed_item}
\item \textit{Compositional structure of real-world objects.} For example, a chair is composed
of a seat, a back-rest and some legs. 
 Having multiple rules for an object helps us capture structural
variations in objects, e.g., a chair can optionally have arm-rests. 
\item \textit{Structure in the 3D arrangement of objects in the scene.} Objects are arranged according to certain preferences in the scene. For example, a monitor is typically placed on a table. Our model allows us to capture such properties.
\item \textit{Non-local properties of objects.} Our model can learn rules about properties of full objects (such as size and shape), each of which comprises several segments. This would be hard (or computationally expensive) to express
in methods based on graphical models \cite{koppula2011semantic}.
\end{packed_item}

\begin{figure}[!t]
 \centering
 \includegraphics[height=0.3\textheight, width=1\textwidth]{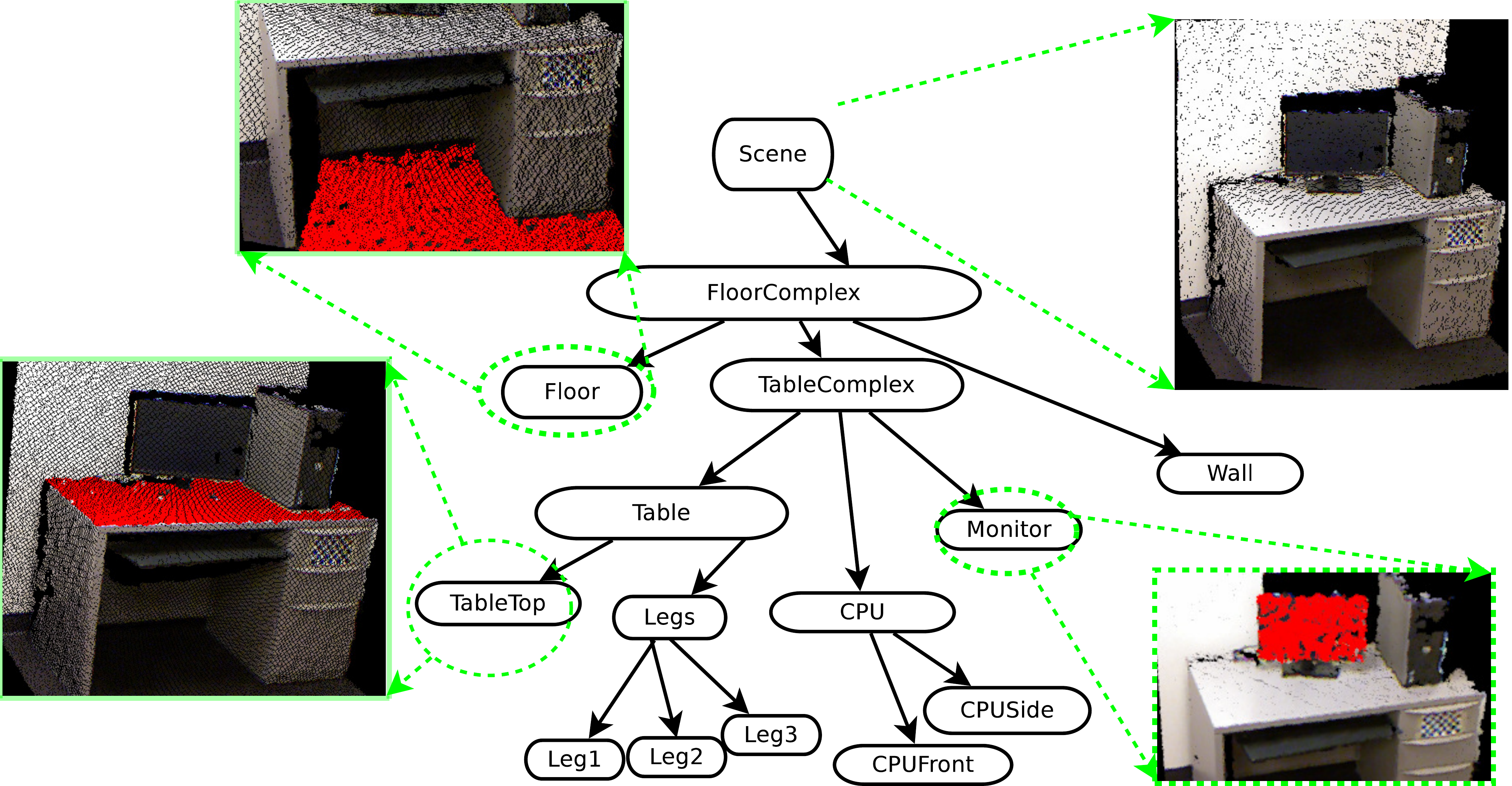}
 \caption{An example of a parse tree for a scene. For an object X, XComplex represents X along with all the objects on top of it. The original tree has been simplified for elucidation.}
\label{fig:IntroEx}
 \end{figure}

Figure \ref{fig:IntroEx} shows an example of a parse tree of a 3D scene in our grammar. 
Intuitively, a grammar defines rules for composing smaller entities to larger entitities.
The rules of our grammar depend on the shape and appearance features of the applicands.
This helps us capture the characteristic configuration in which components of an entity appear in real world, e.g., a table-leg is below a table-top and perpendicular to it. 
Also, since our grammar has rules to combine multiple segments into object-parts, our algorithm does not need perfect segments to begin with.
It can start with segments obtained by an over-segmentation algorithm.
Unlike 1D grammars, exact inference is infeasible in our model.
We therefore use stochastic-beam-search to find the most likely parse tree.



The closest previous work to ours is Zhao et.~al.~\cite{zhu2011}. They used probabilistic grammar to infer the 3D layout of a scene from its 2D image. However, we use 3D data to begin with (i.e., the exact 3D layout is already available). We use grammars to capture the semantic information about 
the 3D scene.
Unlike our work, their parse-tree does not capture semantic information such as the
 labels of objects and their parts.
Our probability model also allows very for efficient training, 
where the training time for our model scales only linearly with the number of rules.
Furthermore, our model is composable: one can combine two separately trained probabilistic sub-grammars and generate a new probabilistic grammar without any additional learning.
This is useful for robotic applications where a robot can download a trained model for 
a new object, plug it into its main grammar and start detecting the object.

We evaluate our method on the Cornell 3D point-cloud dataset \cite{koppula2011semantic}.
In order to quantitatively evaluate our algorithm, we extract the labels of segments from 
the parse-tree and compare the accuracies with their segment-labeling algorithm.
Our method achieves better recall on most objects, but precision was worse in most cases.
More importantly, our model trained within a second as compared to several hours in \cite{koppula2011semantic,anand_koppula:ijrr}---a useful feature when the 
number of objects to be considered increases.





 \section{Related Work\label{sec:related}}

%

There is a huge body of work in the area of scene understanding and object recognition from 
2D images.   Many works address the problem of classifying patches of images into various 
semantic categories, e.g., detecting humans \cite{dalal2005histograms},   detecting \{faces, buildings, trees, etc.\} \cite{csurka2004visual}, global categories such as natural vs. manmade 
scenes \cite{GIST}. All these works work with different input modality and address a different
task, therefore are not directly relevant to ours.

For categorizing 3D data, some works
\cite{Golovinskiy:shape-basedrecognition,Shapovalov2010,Shapovalov2011, xiong:3DSceneAnalysis, Munoz2009:mrf,Anguelov/etal/05,gould:fusion,quigley:high-accuracy,Rusu:ObjectMaps,lai:icra11a, lai:icra11b} address the problem of labeling  the 3D data obtained from LIDAR or Kinect sensors into a few geometric classes (e.g., ground, building, trees, vegetation, etc.). However, these methods consider a `flat' label space and don't capture deeper semantics of the scene in the 2D/3D image.  
At the time of writing this paper, Point-cloud library \cite{pcl} allows one to find basic structures such as planes in the
3D point-clouds, but it does not perform labeling of any complex shapes.
 
Graphical models are often used for capturing the relationships between labels of adjacent regions in 2D/3D images
\cite{Torralba:exploting_context,Hoiem:puttingobjects,Torralba:contextualmodels,HeitzECCV_usingstufftofindthings,Shapovalov2010,Shapovalov2011, xiong:3DSceneAnalysis, Munoz2009:mrf}.
However, such methods lack the ability to combine nodes before labeling them. 
Hence, they cannot capture global properties such as size and shape that can only be 
computed when all the segments belonging to an entity (e.g., object parts) have been combined.

Some works consider capturing composition at some levels in the 2D images. 
For example, 
\cite{felzenszwalb2008discriminatively,girshick2011object,amit2007pop,crandall2005spatial}  
capture the property that an object is a composition of parts.
One can independently run detectors of different objects, but it won't capture relationships between objects when they compose into a scene.
Some works such as \cite{felzenszwalb2008discriminatively} semi-automatically learn the parts required to detect an object in 2D images. 
This is useful when ground-truth labels for objects parts are not available.
However, their algorithm requires that the number of parts be given. More importantly, 
the automatically learnt parts might not be relevant for real applications.
In our work, we made an easy to use GUI for labeling full parse-trees that allows people  to label the full ground-truth parse-tree a pointcloud in about a minute.
Also, in many works, such as  \cite{girshick2011object}, grammar is used to combine parts detected by separate base classifier into an object .
On the other hand, our grammar model coherently handles both these tasks:  detecting parts and combining them into objects.

Felzenszwalb and McAllester \cite{felzenszwalb2007generalized} use a grammar for detecting convex-shapes and salient curves in 2D images.
They present a detailed account of inference in their grammar-model, but do not address the issue of training the same.
In contemporary work, Riemenschneider \cite{shape-grammar-facade} consider
the parsing of building facades in images using properties based on symmetry and low-level
2D detectors such as windows and walls. 
Socher et.~al.~\cite{socher2011parsing} uses recursive neural network to `parse' natural scenes. However, they define a visual tree correct if `all adjacent segments that belong to the same class are merged into one super segment before merges occur with super
segments of different classes.' Therefore, they cannot understand objects at instance level. For example, it would be disastrous for a robot to consider two separate glasses kept nearby as the same object when asked to grab a glass.



To the best of our knowledge, none of the previous works consider parsing 3D point-clouds in a grammar,
and our work is the first one to do so.
Also, unlike many works which use grammars for 2D images, we capture 
composition at almost  all levels of scene-hierarchy.
We consider parts as composition of 3D planar segments, and then consider
consider the whole scene as a composition of objects. 
In this composition, we also capture the property that some objects are observed in characteristic configurations, e.g., a computer monitor is found on a table, and a keyboard in-front-of a monitor.

\section{Scene Grammar}
\label{sec:SceneGrammar}


Our 3D scene grammar a six-tuple $G = (N,\Sigma,P,S,f,g)$.
$\Sigma$ represents the set of terminal nodes. 
In our case, these are the segments generated by over-segmenting the 3D pointcloud given as 
input.  Over segmentation means that an object part could be divided into several small segments.
$N$ are the non-terminal nodes which are obtained by combining terminals and other 
non-terminals---they could be basic units (e.g., Plane), object parts (e.g., TableTop) or 
full objects (e.g., Table).
$S$ is the start symbol which represents the whole scene (the root of the parse-tree). 
$P$ is the set of production-rule schemas that combine a set of terminals and non-terminals 
to form a non-terminal. 

Unlike the usual CFGs, the order of symbols in LHS does not matter. 
The only condition is that a rule can only combine entities close to each other in 3D space.
The input to our algorithm consists of a graph over terminals (segments).
Adjacent segments are connected by an edge in a graph.
This can be viewed as a generalization of the 1D grammars used in NLP, where the 
input (sentence) can be treated as a chain-graph of words.
As in 1D grammars, a non-terminal can only derive a contiguous set of terminals.

\begin{table}[h!]
\caption{Some examples of rules in our 3D scene grammar for the illustrative example in Figure~\ref{fig:IntroEx}. Note that only part of the rules are shown here. The full list of rules can be found in the appendix}
\label{tbl:rules}
\centering
{\small
\begin{tabular}{|l|l|}
\hline
1. & Plane $\rightarrow$ segment\\
2. & Plane $\rightarrow$ Plane segment\\
3. & TableTop $\rightarrow$  Plane\\
\hline
4. & Table  $\rightarrow$ TableTop TableLegLeft TableLegRight\\
5. & Table  $\rightarrow$ TableTop TableLegLeft TableLegRight TableDrawer\\
\hline
6. & TableComplex  $\rightarrow$ Table\\
7 & TableComplex  $\rightarrow$ TableComplex Monitor\\
8 & TableComplex  $\rightarrow$ TableComplex CPU\\
9 & TableComplex  $\rightarrow$ TableComplex Keyboard\\
\hline
10. & S $\rightarrow$ FloorComplex\\
\hline
\end{tabular}
}
\end{table}


We have the following four types of rules in our grammar  that are extracted by our algorithm from the
labeled training set. 
We describe how we extract these rules automatically in the next section.
Table~\ref{tbl:rules} shows an example
of rules for the illustrative example in Figure~\ref{fig:IntroEx}.
\begin{packed_enum}
\item \textit{Segmentation rules.} These rules recursively combine multiple segments into parts of objects. Examples are rules 1-3 in the table.
\item \textit{Object formation rules.}  These rules (e.g., rules 4-5 in the table) combine multiple object parts into one object. Structural variation (e.g., a table having a `table drawer' as well)  is captured by having multiple rules for same object.  Moreover, these structural variants can share the 
same parts thus making full use of the data in training the model. 
Furthermore, in cases of the object being (self-) occluded, we have different rules for 
each set of parts visible respectively in the training dataset.

\item \textit{Object grouping rules.} These rules (e.g., 6-9) capture the relationships between objects. For an object X, we call XComplex as an entity containing X and all objects placed on it. For example, 
Rule 6 forms a TableComplex out of a Table and rules 7-9 recursively add other tabletop objects to it.

\item \textit{Rule to from $S$(goal).}  Rule 10 forms the start symbol of the grammar (goal). Recall 
that $S$ represents the whole scene. In case of indoor scenes, the extracted rule is the 
floor with all objects on top of it (probably via other objects).

\end{packed_enum}



The function $f$ takes a set $s$ of terminals and non-terminals  and returns the corresponding features as a vector of real numbers.
These features contain both the features dependent on individual nodes as well as the relative features between pairs of nodes in $s$.
For example, for a Plane $p$, $f(\{p\})$ contains the normal vector, its centroid position, and area of its convex hull.
For a TableTop $tp$ and a TableLeg $tl$, $f(\{tp,tl\})$ also contains the relative features such as the signed difference between the $z$ coordinates of highest point in $tl$ and lowest point in $tp$. 

The function $g(s,r)$ represents the cost of applying a rule $r$ to a set of nodes $s$. Intuitively, $s$ contains nodes corresponding to the symbols in RHS of the rule $r$. 
We use the negative log probability of applying a rule as it's cost. 
\begin{eqnarray}
g(s,r) = -\log Pr(f(s),r) = -\log Pr(f(s)|r)Pr(r)
\end{eqnarray}
We model $P(f(s)|r)$ with a multivariate Gaussian and $P(r)$ by a categorical  distribution. 
\begin{eqnarray}
g(s,r)&=&-\log(p_r \, Gaussian_{{\mu}_r, {\Sigma}_r}(f(s)) ) 
\end{eqnarray}
Now, following the independence assumptions expressed by the parse tree of the scene (considering
it as a directed graphical model), we have  
the probability of a parse-tree $t$ as:
\begin{eqnarray}
Pr(t)&=& \prod_{n \in NT(t)}{Pr(f(children(n)) , rule(n) )}
\end{eqnarray}
Here, $NT(t)$ denotes the set of non-terminals in the parse-tree. We treat $S$ also as a distinguished non-terminal. Also, $children(n)$ denotes the children of the non-terminal node $n$, and
$rule(n)$ denotes the rule which was applied to $children(n)$ to form node $n$.
The cost of a tree (or subtree) $t$ is:
\begin{eqnarray}
cost(t) = -\log Pr(t) = \sum_{n \in NT(t)}{g(f(children(n)) , rule(n) )}
\end{eqnarray}
Note that $g(s,r)\ge 0$ since it is negative log of a probability.
Therefore, we have an important property that \emph{the cost of a node cannot be less than the
cost of any of its children}. This will be useful later for the inference algorithms.

Furthermore, also note that
we can trivially compose two probabilistic sub-grammars to form a new probabilistic 
grammar with only performing incremental learning.
Suppose we want to add a new  table-top object (say a laptop) to our grammar.
We can just add the rules for forming laptop to 
$P$ and the corresponding (unique) symbol names to $N$.
Because of the independence assumptions, the probability function for rules to form a laptop would remain unchanged when added to the grammar.
I.e., we need to add the rule $r_{new}$: \{TableComplex  $\rightarrow$ TableComplex laptop\},
and the $Pr(f({\rm laptop}, {\rm TableComplex}),r_{new})$.

\section{Learning Algorithm}
Our learning algorithm comprises of four steps: rule extraction, rule splitting, learning
the rule dependent probability, and learning a rule's prior probability.


\textbf{Rule Extraction.}
Our algorithm for rule extraction maintains a working set of rules. It
is first initialized to an empty set, and then it traverses all the ground-truth labeled parse trees.
Whenever it encounters a non-leaf node of type $T_p$ with children of types $T_{c_1},...,T_{c_n} $, it adds the following rule to working set:
$T_p$ $\rightarrow$ $T_{c_1}$  $T_{c_2}$ ..  $T_{c_n}$. 
Recall that the order of symbols in RHS of a rule is irrelevant.

\textbf{Rule Splitting.}
For efficient parsing, we make sure each production rule has a maximum of two symbols in the RHS. We do this by introducing intermediate nonterminals. For example, conside the rule:
\{CPU $\rightarrow$ CPUFront CPUTop CPUSide\}.
We can now introduce an intermediate non-terminals CPU\_FS. The new rules would
then be: \{CPU\_FS $\rightarrow$ CPUFront CPUSide\}, and \{CPU $\rightarrow$ CPU\_FS CPUTop\}.


\textbf{Learning the rule dependent probability.}
Recall that the probability of applying rule $r$ to set of nodes $s$ is:
\begin{equation}
P(f(s),r) =P(f(s)|r)P(r)
\end{equation}
Here, $P(f(s)|r)$ is the component which depends on the features of of the $s$, the set of applicands of rule $r$. $f(s)$ represents the features of $s$.  We model $P(f(s)|r)$ with one of the following distributions: multi-variate Gaussian or an exponential. 

When the nodes $s$ are 3D entities, we use a multi-variate Gaussian. The parameters ${\mu}_r, {\Sigma}_r$ can be  computed by fitting a Gaussian to the set of feature vectors obtained from all the instances where the rule $r$ was applied.

When the entities involve planar surfaces (e.g., rules 1-2 in Table~\ref{tbl:rules}), we use a 
Laplacian distribution. Specifically we have 
$Pr \propto \exp(-d)$, where $d$ is the sum of squared distance of the points 
in the newly formed plane to the best-fit plane.\footnote{For efficiency we do not store the spanned set of points for each non-terminal.  We simply maintain the $3\times 3$ scatter matrix for the points spanned by a non-terminal and use it's eigen decomposition to compute the best-fit plane and $d$.}
For the special case of rule involving $S$ (root), we always have 
$\exp(-kn)$, where $k$ is a positive constant and $n$ is the number of terminals not spanned by the newly formed goal node.

\textbf{Learning a rule's prior probability.}
We model $P(r)$ by a categorical distribution. This is computed as follows.
For a non-terminal $A$, let there be $n$ rules with $A$ in LHS : $r_1$ ... $r_n$. 
Then, $P(r_i)$ is the fraction of instances in the training set where $A$ was formed 
using rule $r_i$.

\subsection{Features}
Our features are of two types: node and edge features.  Each type of feature comprises
shape as well as appearance features. They are similar to the ones in \cite{koppula2011semantic}. 
In the context of our grammar, if the RHS of a rule has only one symbol, the the node features are used.
If it has more than one symbols, edge features between each pair is also appended to the 
feature vector.
Whenever we have an intermediate node (e.g., CPU\_FS), we expand it into all the non-intermediate
parts of the object (CPUFront and CPUSide). Thus, 
$f(\{CPU\_FS, CPUTop\})$ would contain the relative features between all the three parts (CPUTop, CPUFront and CPUSide).

\section{Computing Predictions}
The generalization from 1D to 3D grammars has adverse implications on the time-complexity of algorithms for finding the most-likely parse-tree. The dynamic-programming (DP) 
based algorithms for 1D grammars finds the optimal parse-tree in polynomial time. 
This is because, the number of sets of contiguous nodes (terminals) in a chain graph containing $n$ nodes is only $O(n^2)$. However,
the same is not true for 3D adjacency graphs in general. Even a 2D grid graph containing $n^2$ nodes can be shown to have at least $2^n$ distinct sets of contiguous nodes. So, a DP based solution 
is nearly hopeless (unless the adjacency graph is very sparse).

We can however, do better by exploiting a special property of our cost function. As mentioned 
in Section~\ref{sec:SceneGrammar}, the cost of a sub-tree of a node is never less that cost of 
any of its children sub-trees.
Hence, we can use the KLD (Knuth Lightest Derivation) algorithm  \cite{felzenszwalb2007generalized}) to find a least cost (most likely) parse tree. 
KLD is a generalization of the Dikstra shortest path algorithm for finding 
`shortest` derivations in a grammar.
Unlike the DP-based algorithm, KLD will only create the nodes which cost less that 
the most likely parse-tree.
In our experients, the KLD algorithm achieves very high accuracies, but it 
does not terminate soon enough on many scenes. 

For fast inference, we also used a stochastic beam search algorithm \cite{aima} to 
deal with the cases where KLD does not terminate fast enough. The states in the beam 
are forests. There are a collection of partial parse-trees.
We initialize the beam by a single state containing an isolated node for each terminal.
The next state of a forest is obtained a applying rule to some subtrees in the forest.
The probability of choosing a next state is proportional to the probability of the rule application 
required to get that state. The algorithm keeps on generating next states till no rule can 
be applied to any forest in the beam. 
This algorithm is fast in execution time, but
is not guaranteed to produce an optimal parse-tree.

\vspace*{\sectionReduceTop}
   \section{Experiments\label{sec:experiments}}

   \subsection{Data}
 
 We evaluate our method on the Cornell 3D pointcloud dataset \cite{anand_koppula:ijrr},
 which comprises  84 single-view pointclouds, for the classes we had more
 than 10 instances.
We first over-segment the 3D scene (as described earlier)
to obtain the terminals of the parse-tree.
We then find the adjacency graph over terminals.
Two terminals are connected by an edge if they are less than 0.05 metre apart, 
or less than 0.5 metre apart if the region between segments is occluded.
For training our model, we manually annotated ground-truth parse-trees for each scene.
To do this, we  wrote an easy to use interactive parse-tree labeler 
which is now released as an open-source ROS package.

 \begin{figure}[t!]
 \centering
 \includegraphics[width=.47\linewidth,height=2.1in]{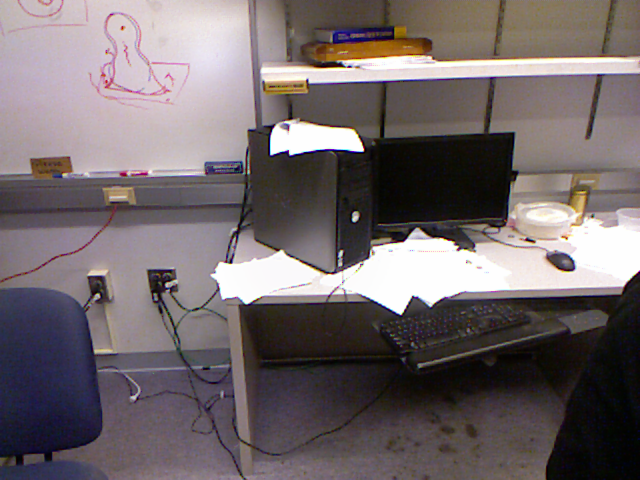}
 \hskip .2in
\includegraphics[width=.47\linewidth,height=2.1in]{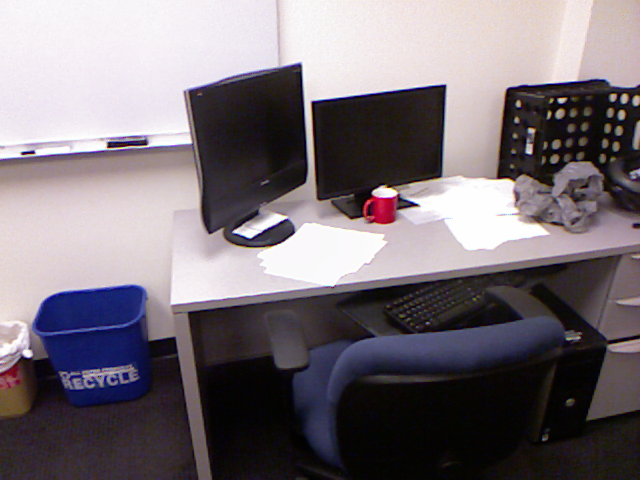}
 \includegraphics[width=1\linewidth,height=2.2in]{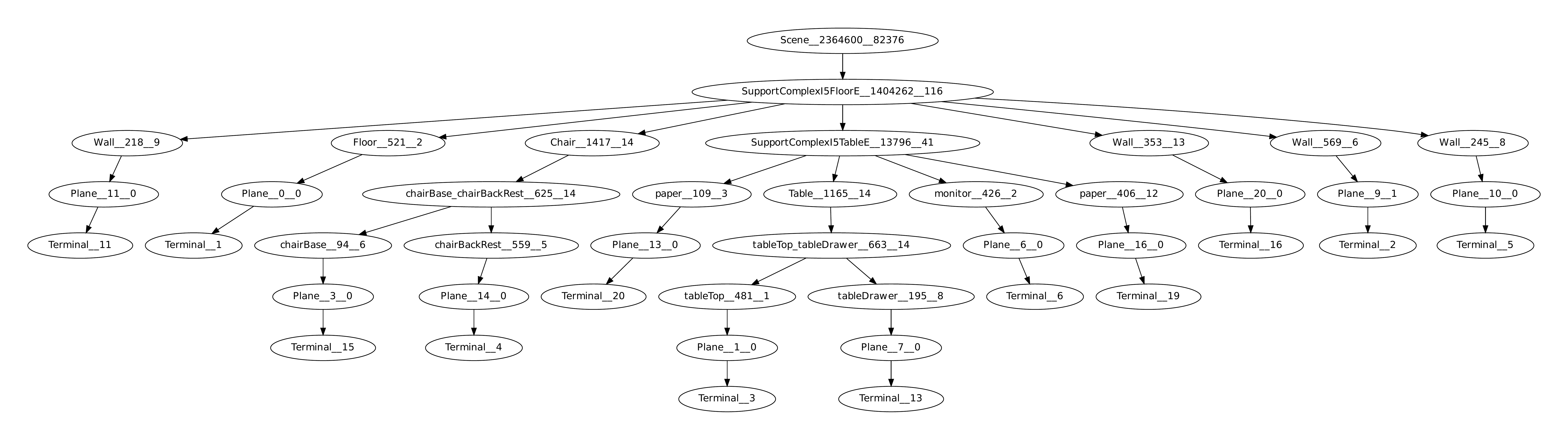}\\
\includegraphics[width=1\linewidth,height=2.4in]{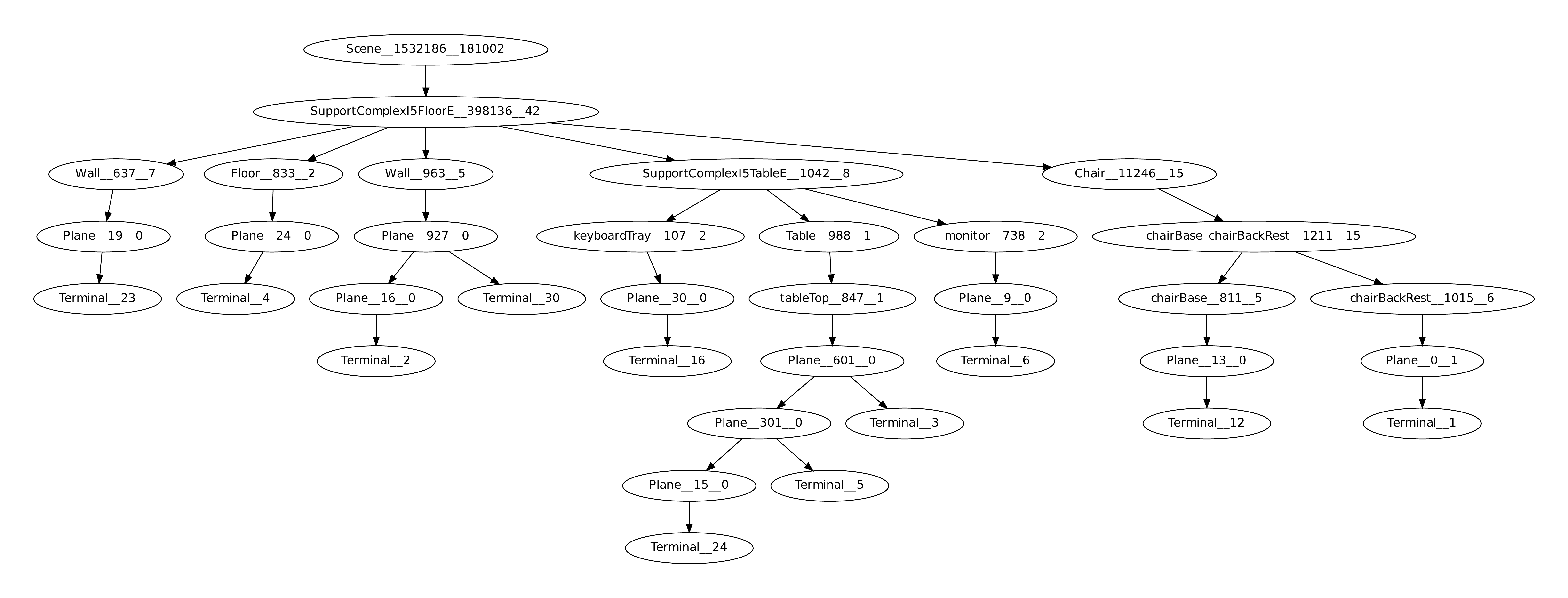} \\
 \caption{Top row shows two office scenes. Middle row shows the inferred parse tree for the scene
 on the left, and bottom row shows the inferred parse tree for the scene on the right. This figure is in a vector graphics format and retains quality on zooming. Please view it at high zoom settings. }
\vspace*{\captionReduceBot}
\label{fig:parseTree}
\end{figure}


   \subsection{Results}
   
We evaluate our method in two ways: First, on segment labeling, computing
the precision and recall in labeling at the segment level. Second, on the quality of the
parse tree inferred.

The segment labeling evaluation follows same 
procedure as in \cite{anand_koppula:ijrr}.   In order to do this, we extract the labels of 
segments from the parse-tree and
compare the accuracies with their segment-labeling algorithm \cite{anand_koppula:ijrr}.
   We used four-fold cross-validation to evaluate our model. 
   Table \ref{tbl:overall_result} shows the object-wise precision and recall for each object part. 
   We see that it achieves higher recall on many objects, e.g., tableTop, chairBackRest, but precision 
   is sometimes low for many classes. We observe that it often confuses object parts with one another, such as chairBack as chairBackRest.
   
   
   However, the goal of our method is to compute a parse tree, and labeling a segment
   is only part of the task.  One can label a segment correctly, but still associate it with incorrect
   object.  We   show  some examples of parse-trees inferred by our algorithm
   in  Figures \ref{fig:parseTree}. For the mid-left scene, it parsed the chair, monitor and walls perfectly.
   There was a minor mistake in parsing the table (tableLeg was mistaken as tableDrawer). However, CPU parts were  mistaken as walls. The paper on top of CPU was completely missed.
      For the mid-right scene, table, wall, keyboardTray and the monitor in RHS were parse correctly. There was a minor mistake in parsing the chair (chairBack was confused as chairBackRest).  The LHS monitor was missed because it had very few points in the pointcloud (pointclouds suffer from the problem of missing points---which happens when the surface is reflective for example).


Our model trains in a second, as compared to several hours in methods based on graphical models \cite{anand_koppula:ijrr}, which model $O(n^2)$ terms for $n$ objects.
Our method can thus potentially scale to 
modeling internet-scale collection of objects.


\begin{table*}[tb!]
\caption{{\small {\bf Labeling experiment statistics.} This table shows object-wise accuracies for labels extracted out of parse-trees}.}
 \label{tbl:overall_result}
 \centering
{\small
\begin{tabular}{l|c|cccccccccccc|c}
\whline{1.1pt}
method &  metric & \begin{sideways} Wall  \end{sideways}& \begin{sideways}Floor\end{sideways} & \begin{sideways} tableTop \end{sideways} & \begin{sideways} chairBackRest \end{sideways} & \begin{sideways} CPUFront \end{sideways} & \begin{sideways} monitor \end{sideways} & \begin{sideways} paper \end{sideways} & \begin{sideways} tableLeg \end{sideways} & \begin{sideways} keyboard \end{sideways} & \begin{sideways} chairBase \end{sideways} & \begin{sideways} tableDrawer \end{sideways} & \begin{sideways} chairBack \end{sideways} 
& \begin{sideways} macro avg. \end{sideways} \\
\whline{1.1pt}
\multirow{2}{*}{Anand et.~al.~\cite{anand_koppula:ijrr}} & recall & 70 & 81 & 62 & 44 & 23 & 46 & 12 & 33 & 7 & 51 & 43 & 39 & 43\\
 & precision & 93 & 92 & 85 & 80 & 100 & 85 & 100 & 83 & 50 & 95 & 88 & 94  & 87\\
\hline  
  \multirow{2}{*}{Our method}  & recall & 83 & 85 & 73 & 58 & 24 & 30 & 16 & 32 & 17 & 49 & 32 & 0   & 42 \\
 & precision & 64 & 100 & 62 & 25 & 90 & 49 & 50 & 28 & 50 & 59 & 31 & 0   & 51\\
\whline{1.1pt}
\end{tabular}
}
\end{table*}

\section{Conclusion}

In this paper, we presented a method to parse the 3D point-cloud data,
obtained from inexpensive RGB-D sensors, using a scene grammar.
Our grammar allowed us to capture the compositional nature of the
real-world objects as composed of parts, as well as the relative 3D arrangements of
the objects in the scene. Furthermore, our method is robust to variations
in the compositional rules (e.g., some chairs do not have arm-rests, or a table
leg may be occluded) as well as to variations in the shape and appearance
of the objects. Expressing the 3D scene parsing as a compositional grammar 
also allowed us to build a incremental learning algorithm where two different
sub-grammars can be easily composed. 
Although, our results do not beat the state-of-the-art for scene-labeling,
we think this is an important first attempt at a harder problem.

{  
 \bibliographystyle{abbrv}
\bibliography{references}
}

\pagebreak
\section{Appendix: list of rules}
\begin{verbatim}
pillarLeft --> Plane
printerTop --> Plane
pillarRight --> Plane
printerSide --> Plane
tableDrawer --> Plane
chairArmRest --> Plane
keyboardTray --> Plane
printerFront --> Plane
sofaBackRest --> Plane
chairBackRest --> Plane
FloorComplex --> FloorComplex , TableComplex
FloorComplex --> FloorComplex , door
FloorComplex --> FloorComplex , sofa
FloorComplex --> FloorComplex , Wall
FloorComplex --> FloorComplex , Chair
FloorComplex --> FloorComplex , Table
FloorComplex --> FloorComplex , ACVent
FloorComplex --> FloorComplex , pillar
FloorComplex --> FloorComplex , printer
FloorComplex --> Floor
TableComplex --> TableComplex , keyboardTray
TableComplex --> TableComplex , CPU
TableComplex --> TableComplex , paper
TableComplex --> TableComplex , monitor
TableComplex --> TableComplex , printer
TableComplex --> TableComplex , keyboard
TableComplex --> Table
CPUTop_CPUFront --> CPUTop , CPUFront
CPUSide_CPUFront --> CPUSide , CPUFront
tableLeg_tableTop --> tableLeg , tableTop
tableTop_tableDrawer --> tableTop , tableDrawer
sofaBackRest_sofaBase --> sofaBackRest , sofaBase
pillarRight_pillarLeft --> pillarRight , pillarLeft
chairBase_chairBackRest --> chairBase , chairBackRest
CPUSide_CPUFront_CPUTop --> CPUSide_CPUFront , CPUTop
printerFront_printerTop --> printerFront , printerTop
tableTop_tableDrawer_tableLeg --> tableTop_tableDrawer , tableLeg
chairBase_chairBackRest_chairBack --> chairBase_chairBackRest , chairBack
tableTop_tableDrawer_keyboardTray --> tableTop_tableDrawer , keyboardTray
printerFront_printerTop_printerSide --> printerFront_printerTop , printerSide
chairBase_chairBackRest_chairArmRest --> chairBase_chairBackRest , chairArmRest
CPU --> CPUTop_CPUFront
CPU --> CPUSide_CPUFront
CPU --> CPUSide_CPUFront_CPUTop
tableTop_tableDrawer_tableLeg_keyboardTray --> tableTop_tableDrawer_tableLeg 
, keyboardTray

door --> Plane
Wall --> Plane
tableTop_tableDrawer_tableLeg_keyboardTray_tableBack --> 
tableTop_tableDrawer_tableLeg_keyboardTray , tableBack

Chair --> chairBase_chairBackRest
Chair --> chairBase_chairBackRest_chairBack
Chair --> chairBase_chairBackRest_chairArmRest
Floor --> Plane
paper --> Plane
Table --> tableLeg_tableTop
Table --> tableTop_tableDrawer_tableLeg
Table --> tableTop_tableDrawer_keyboardTray
Table --> tableTop_tableDrawer_tableLeg_keyboardTray_tableBack
Table --> tableTop
ACVent --> Plane
CPUTop --> Plane
pillar --> pillarRight_pillarLeft
CPUSide --> Plane
monitor --> Plane
printer --> printerFront_printerTop
CPUFront --> Plane
keyboard --> Plane
sofaBase --> Plane
tableLeg --> Plane
tableTop --> Plane
chairBack --> Plane
chairBase --> Plane
tableBack --> Plane
\end{verbatim}

\end{document}